\documentclass{article}
\usepackage{spconf,amsmath,graphicx}
\usepackage{multirow,soul,xcolor,boldline,hhline,hyperref, enumitem,wrapfig, cite}
\usepackage{pifont}
\newcommand{\cmark}{\ding{51}}%

\newcommand{\tablestyle}[2]{\setlength{\tabcolsep}{#1}\renewcommand{\arraystretch}{#2}\centering\footnotesize}

\newif\ifarxiv
\arxivtrue

\newif\ifdraft
\drafttrue
\draftfalse

\definecolor{mygray}{gray}{0.7}
\definecolor{dkgreen}{RGB}{0,179,36}
\definecolor{dkorange}{RGB}{230,115,0}

\newcommand{\suwon}[1]{\ifdraft \textcolor{cyan}{[#1 --Suwon]} \fi}
\newcommand{\ap}[1]{\ifdraft \textcolor{dkgreen}{[AP: #1]} \fi}
\newcommand{\kl}[1]{\ifdraft \textcolor{blue}{[KL: #1]} \fi}

\newcommand{\pb}[1]{\ifdraft \textcolor{brown}{[PB: #1]} \fi}

\ifdraft
\newcommand{\apedit}[1]{\textcolor{dkorange}{{#1}}}
\newcommand{\apremove}[1]{\textcolor{dkorange}{{\st{#1}}}}
\newcommand{\kledit}[1]{\textcolor{red}{{#1}}}
\newcommand{\klremove}[1]{\textcolor{blue}{{\st{#1}}}}

\newcommand{\khremove}[1]{\textcolor{magenta}{{\st{#1}}}}
\newcommand{\suwondel}[1]{\textcolor{cyan}{{\st{#1}}}}
\newcommand{\suwonadd}[1]{\textcolor{cyan}{{#1}}}
\else
\newcommand{\apedit}[1]{\textcolor{black}{{#1}}}
\newcommand{\kledit}[1]{\textcolor{black}{{#1}}}
\newcommand{\apremove}[1]{{}}
\newcommand{\klremove}[1]{{}}

\newcommand{\khremove}[1]{{}}
\newcommand{\suwonadd}[1]{\textcolor{black}{{#1}}}
\newcommand{\suwondel}[1]{{}}
\fi

\title{SLUE: New Benchmark Tasks \\ for Spoken Language Understanding Evaluation on Natural Speech}

\name{Suwon Shon$^1$, Ankita Pasad$^{2^*}$\thanks{$^*$Work done during an internship at ASAPP}, Felix Wu$^1$, Pablo Brusco$^1$, Yoav Artzi$^{1,3}$, Karen Livescu$^{2}$, Kyu J. Han$^1$}
\address{
  $^1$ASAPP \ \ \ \ \   $^2$Toyota Technological Institute at Chicago \ \ \ \ \  $^3$Cornell University}

\begin{document}
\ninept
\maketitle

\begin{abstract}
\vspace{-1mm}
Progress in speech processing has been facilitated by shared datasets and benchmarks. Historically these have focused on automatic speech recognition (ASR), speaker identification, or other lower-level tasks. Interest has been growing in higher-level spoken language understanding tasks, including using end-to-end models, but there are fewer annotated datasets for such tasks. At the same time, recent work shows the possibility of pre-training generic representations and then fine-tuning for several tasks using relatively little labeled data. We propose to create a suite of benchmark tasks for Spoken Language Understanding Evaluation (SLUE) consisting of limited-size labeled training sets and corresponding evaluation sets. This resource would allow the research community to track progress, evaluate pre-trained representations for higher-level tasks, and study open questions such as the utility of pipeline versus end-to-end approaches.
We present the first phase of the SLUE benchmark suite, consisting of named entity recognition, sentiment analysis, and ASR on the corresponding datasets. We focus on naturally produced (not read or synthesized) speech, and freely available datasets. We provide new transcriptions and annotations on subsets of the VoxCeleb and VoxPopuli datasets, evaluation metrics and results for baseline models, and an open-source toolkit to reproduce the baselines and evaluate new models.


\noindent

\end{abstract}

\begin{keywords}
spoken language understanding, benchmark, pre-training, named entity recognition, sentiment analysis
\end{keywords}
\section{Introduction}
\label{sec:intro}
\vspace{-2mm}
\ap{I think we should also mention somewhere in the abstract and/or intro the language(s) we'll be focusing on with this benchmark. Either that this is English or that we plan on extending it to other languages (if we do).}\kl{agreed}
Progress on speech processing has benefited from shared data sets and benchmarks.
For tasks with plentiful shared resources, 
such as ASR, we now have high-performing and commercially feasible systems, at least for high-resource languages.  On the other hand, “higher-level” spoken language understanding (SLU) tasks have received less attention and resources.  There are numerous tasks, at varying linguistic levels, that have been benchmarked extensively for text input by the natural language processing (NLP) community---named entity recognition, parsing, sentiment analysis, entailment, summarization, and so on---
but they have not been as thoroughly addressed for speech input. 
Better benchmarks would allow the research community to address open research questions about SLU, such as which tasks can be addressed well by pipeline ASR+NLP approaches and which ones benefit from having direct access to the input speech; and, for the latter kind of tasks, how to best extract the needed speech information.

One challenge for advancing SLU is that it is costly and time-consuming to collect large amounts of labeled speech data for all of these tasks. On the other hand, recent developments suggest that it is possible to pre-train general speech representations that can then be fine-tuned for a variety of tasks, given only a small amount of labeled data~\cite{chung2019unsupervised,pascual2019learning,yang2021superb}. It is this combination of factors---the need for more benchmark SLU tasks, along with advances in pre-trained representations---that motivate our work. Specifically, the goals of our work are to (1) {\bf track research progress} on multiple SLU tasks, (2) {\bf facilitate the development of pre-trained representations} by providing fine-tuning and eval sets
for a variety of SLU tasks, and (3) \textbf{foster the open exchange of research} by focusing on freely available datasets that all academic and industrial groups can easily use.

Several other recent efforts have produced multi-task benchmarks, either combining existing data sets and tasks~\cite{yang2021superb} or generating read speech data from existing NLP task data~\cite{feng2021asrglue}.  In this work, we contribute the following: (i) New \textbf{annotation of publicly available, natural speech data} for training and evaluation on new tasks, specifically {\bf named entity recognition} (NER) and {\bf sentiment analysis} (SA), as well as new text transcriptions for training and evaluating ASR systems on the same data.  
\kl{moved footnote placement (throughout the paper).  footnotes go after commas/periods/semicolons, not before (the only punctuation where footnotes go before are dashes)}
(ii) A \textbf{benchmark suite} including a toolkit for baseline models and evaluation,\footnote{\href{https://github.com/asappresearch/slue-toolkit}{https://github.com/asappresearch/slue-toolkit}} the annotated data, web site, and leaderboard.\footnote{https://asappresearch.github.io/slue-toolkit/leaderboard.html}
(iii) \textbf{A variety of baseline models} to measure the state of existing models on the\apedit{se} new tasks.

\vspace{-4mm}
\section{Related work}
\label{sec:related}
\vspace{-1mm}

{\bf Related work on pre-trained representations.}
In recent NLP research, self-supervised language models trained on unlabeled text~\cite{devlin2018bert,liu2019roberta,he2020deberta} have come to be used as more or less universal representations for a large range of downstream tasks.  Self-supervised speech representations~\cite{schneider2019wav2vec,jiang2019improving,ling2020deep,baevski2020wav2vec,wang2020unsupervised,ravanelli2020multi,hsu2021hubert,wu2021performance} have also become quite successful for a number of tasks, but it is not yet clear how generally applicable any one model is to a variety of downstream tasks.  Most pre-trained models are evaluated on a single task such as ASR~\cite{schneider2019wav2vec,jiang2019improving,baevski2020wav2vec,wang2020unsupervised,ravanelli2020multi,hsu2021hubert,wu2021performance}, speaker recognition~\cite{ravanelli2018learning,huh2020augmentation}, slot filling~\cite{wang2020large}, or emotion recognition~\cite{li2021contrastive,eskimez2018unsupervised}.  A few studies have focused on fine-tuning a pre-trained model for multiple downstream tasks, but mostly low-level tasks such as phoneme recognition, speaker identification, keyword spotting, and emotion recognition~\cite{oord2018representation,chung2019unsupervised,pascual2019learning,liu2020mockingjay,chi2021audio} (with some exceptions; e.g., Liu et al.~\cite{liu2020mockingjay} evaluate on sentiment analysis).

\noindent{\bf Benchmark suites and tasks.} 
SUPERB~\cite{yang2021superb} is a recently created benchmark collection of pre-existing speech processing tasks, designed to compare the applicability of pre-trained models on a variety of tasks, most of which are lower-level speech tasks.  It also includes two SLU tasks, intent classification (IC) and slot filling (SF). However, the dataset used for the IC task~\cite{lugosch2019speech} is one where many recent models achieve nearly perfect performance, and the dataset for SF~\cite{coucke2018snips} \kledit{consists of artificial (synthesized) rather than natural speech.}\klremove{is artificial speech synthesized using Amazon Poly with limited variety of voice.}
ASR-GLUE~\cite{feng2021asrglue} is another recently proposed benchmark, which consists of tasks from GLUE~\cite{wang2018glue}, the popular NLP benchmark, converted to speech by having six speakers dictate the text. Although it includes many more SLU tasks, the limited set of speakers and read speech format makes it less natural.  In addition, no in-domain training speech data is available for the ASR-GLUE tasks.  

In addition to these benchmark suites, there are well-known SLU tasks such as ATIS~\cite{hemphill1990atis} and the Switchboard NXT tasks~\cite{calhoun2010nxt}. \suwondel{However, these have some drawbacks the audio is too clean the data is not free or has licensing constraints.}\suwonadd{However, these data sets have licensing constraints, and in the case of ATIS, it is easy to achieve near-perfect performance~\cite{lai2021semi} since \kledit{the} audio is too clean.} \kl{is it just that the audio is too clean?  If the audio were clean but the task were more complex, it would still be an interesting task.  We could just remove "since the audio is too clean".} 
For the tasks we consider in this paper---NER and sentiment---other previously released datasets have clear limitations: 1) sentiment analysis datasets are either scripted~\cite{busso2008iemocap}, single-speaker monologues~\cite{zadeh2018multimodal}, or not freely available~\cite{chen2020large, martinez2020msp}. 
2) the available NER speech dataset\kledit{s}~\cite{yadav2020end} we are aware of are a mix of dictated text and TED talks, with a somewhat biased set of annotations since they are filtered by a text-based NER tagger; and data that is not freely available~\cite{cohn2019audio}. 

\vspace{-2mm}
\section{SLUE Benchmark}
\vspace{-3mm}

\label{sec:tasks}
\subsection{Tasks}
\textbf{ASR:} 
Although it is not a SLU task, we include ASR because it can help analyze performance on downstream SLU tasks on the same domain, and because pipeline approaches depend on ASR outputs.
ASR is evaluated using word error rate (WER).

\noindent\textbf{NER:}
Named entity recognition involves detecting the named entities and their tags (types) in a given sentence. Named entities are phrases, 
often (but not always) consisting of proper nouns, that refer to distinct entities such as a person, location, organization, numerical value, etc. NER is relevant to downstream tasks like de-identification~\cite{cohn2019audio} and coreference resolution~\cite{durrett2014joint}.

Similarly to previous work~\cite{ghannay2018end, yadav2020end}, we evaluate performance using micro-averaged F1 scores on two aspects of the output. F1 score is the harmonic mean of precision and recall. 
The first score, referred to as F1, evaluates an unordered list of named entity phrase and tag pairs predicted for each sentence. 
The second score, label-F1, considers only the tag predictions. Label-F1 is useful to understand model accuracy despite the possible misspelling and segmentation errors in speech-to-text conversion.

\noindent\textbf{SA:}
Sentiment analysis refers to classifying a given speech segment as having negative, neutral, or positive sentiment.
It is a higher-level task since the semantic content is also very important. For example, negative sentiment can be expressed by disparagement, sarcasm, doubt, suspicion, frustration, etc.~\cite{mohammad2016practical}. We evaluate SA using  macro-averaged (unweighted) recall and F1 scores, similarly to previous work~\cite{lu2020speech}.

\vspace{-2mm}
\subsection{Data and annotation}
\label{sec:annotation}
The data statistics are shown in Tab.~\ref{tab:tasks_description}. The fine-tune and dev sets are intended for training and validation, while the test set is held out for final evaluation. 
We hired several expert transcribers and annotators from a third-party vendor and an internal annotation team.
\begin{table*}[h]
\centering
\resizebox{1.0\linewidth}{!}{%
\begin{tabular}{l|rrr|l|l|l|l}
\hlineB{2}
\multirow{2}{*}{Corpus} & \multicolumn{3}{c|}{Size - utts (hour)} & \multirow{2}{*}{Tasks} & \multirow{2}{*}{Speech type} & \multirow{2}{*}{Source domain} & \multirow{2}{*}{License} \\ \cline{2-4}
 & \multicolumn{1}{c}{Fine-tune} & \multicolumn{1}{c}{Dev} & \multicolumn{1}{c|}{Test} &  & & &\\ \hlineB{2}
SLUE-VoxPopuli  & 5,000 (14.5)  & 1,753 (5.0)  & 1,842 (4.9) & ASR, NER & Scripted & European parliament & CC0\\ \hline
SLUE-VoxCeleb & 5,777 (12.8) & 1,454 (3.2) & 3,553 (7.8)  & ASR, SA & Conversational & Broadcasting (YouTube) & CCBY 4.0\\ \hlineB{2}
\end{tabular}%
}
\vspace*{-3mm}
\caption{Corpus statistics and task descriptions of SLUE benchmark v0.2.
}
\label{tab:tasks_description}
\vspace{-4mm}
\end{table*}


\vspace{-2mm}

\subsubsection{VoxPopuli dataset: NER annotation}
\vspace{-1mm}

\begin{table}[]
\centering
\resizebox{0.95\linewidth}{!}{%
\begin{tabular}{c|l|r|r|r}
\hlineB{2}
\multicolumn{1}{c|}{\multirow{2}{*}{\begin{tabular}[c]{@{}c@{}}Combined\\ label\end{tabular}}} & 
\multicolumn{1}{c|}{\multirow{2}{*}{\begin{tabular}[c]{@{}c@{}}Raw label\\ (ontonotes5)\end{tabular}}} & 
\multicolumn{3}{c}{\# of NER phrases} \\
& & Fine-tune & Dev & Test \\\hline
PLACE & GPE, LOC & 2012 & 642 & 731 \\[3pt]
QUANT & \begin{tabular}[l]{@{}l@{}}CARDINAL, MONEY,\\ ORDINAL, PERCENT,\\ QUANTITY\end{tabular} & 923 & 327 & 246 \\[0.5cm]
ORG & ORG & 864 & 259 & 273 \\[3pt]
WHEN & DATE, TIME & 762 & 260 & 186 \\[3pt]
NORP & NORP & 647 & 220 & 348 \\[3pt]
PERSON & PERSON & 272 & 51 & 81 \\[3pt]
LAW & LAW & 250 & 60 & 96 \\[3pt]
\hline
\hlineB{2}
\end{tabular}%
}
\vspace*{-1mm}
\caption{SLUE-VoxPopuli v0.2 NER label statistics.}
\label{tab:voxpopuli_stats}
\vspace{0mm}
\end{table}

VoxPopuli~\cite{wang2021voxpopuli} is a large multilingual speech corpus consisting of European Parliament event recordings with audio, transcripts and timestamps from the official Parliament website.
By the nature of the source, the spoken data includes abundant named entities making it an ideal choice for NER. We use the English subset of the data with transcriptions. We retain the canonical splits provided in the official repository\footnote{\href{https://github.com/facebookresearch/voxpopuli}{https://github.com/facebookresearch/voxpopuli}} for the dev and test sets. For the fine-tune set, we sample about 15 hours from the official train set.

For annotation, we follow the OntoNotes Release 5.0~\cite{hovy2006ontonotes} guidelines and entity labels. The label-wise counts in the annotated data are reported in Tab.~\ref{tab:voxpopuli_stats}. As the domain of OntoNotes 5 is slightly different from VoxPopuli, for evaluation, we combine similar categories
and discard the rare ones, resulting in 7 categories. Raw labels before combining are still included in the dataset.
We hired 4 annotators and all annotation was done on text transcripts.

To estimate human performance, we obtain a second pass of annotations for the test set. The second pass achieved an micro-averaged F1 score of 0.79 when evaluated against the first pass. The disagreement between the two passes can either be classified as a mismatch in the detection of the entity phrase (missed/over/partial detection) or a mismatch in the label when they agree on the entity phrase (mislabel). We see that 88\% of these disagreements were detection errors and on a closer look at the data, we notice certain recurring systematic differences in the two passes leading to a majority of these errors.  We will collect more passes with  updated guidelines to account for these objective disagreements, in order to report a more robust human performance. 


\vspace{-2mm}
\subsubsection{VoxCeleb dataset: Transcription and Sentiment Annotation}
\vspace{-1mm}

\begin{table}[]
\centering
\resizebox{0.7\linewidth}{!}{%
\begin{tabular}{c|rrr}
\hlineB{2}
\multirow{2}{*}{Sentiment label} & \multicolumn{3}{c}{\begin{tabular}[c]{@{}c@{}}Sets\end{tabular}}  \\ \cline{2-4} 
 & \begin{tabular}[c]{@{}l@{}}Fine-tune\end{tabular} & \begin{tabular}[c]{@{}l@{}}Dev\end{tabular} & \begin{tabular}[c]{@{}c@{}}Test\end{tabular} \\ \hline
Negative & 227 & 51 & 168 \\ 
Neutral & 4,223 & 1,124 & 2,521 \\ 
Positive & 1,279 & 262 & 737 \\ 
Mixed & 48 & 3 & 23 \\ \
Disagreement & - & 14 & 104 \\ \hline \hline
Total & 5,777 & 1,454 & 3,553 \\ \hlineB{2}
\end{tabular}
}
\vspace*{-1mm}
\caption{SLUE-VoxCeleb v0.2 Sentiment label distribution.
}
\label{tab:voxceleb_stats}
\vspace{-4mm}
\end{table}
The VoxCeleb \cite{nagrani2017voxceleb} dataset consists of unrestricted single-sided conversation voice snippets for several thousand people extracted from YouTube videos and was originally created for speaker recognition. 
We provide transcripts and sentiment annotations for the English portion of the official VoxCeleb1 test set (35 out of 40 speakers), and also for a 15-hour subset of the official VoxCeleb1 dev set, which we divide into our fine-tune set (120 speakers) and dev set (20 speakers).
We exclude utterances containing slurs, and trim partial words from the ends of audio segments using a forced aligner (as reflected in the timestamps that we provide).

Sentiment labels were obtained by asking annotators to listen to each individual utterance within each conversation maintaining the order of appearance. This dataset contains utterances only for the interviewee and not the interviewer, so our annotators had to imagine the rest of the conversation. 
Following~\cite{chen2020large}, the annotators labeled an utterance as
\textbf{positive} when the speaker showed signs of happiness and positive attitude, such as laughing or smiling or using positive words (encouragement, joy, etc.); as \textbf{negative} when the speaker showed signs of negative emotions such as raising voice in anger, being dismissive, or using negative words (including disparagement, doubt/questioning/suspicion, sarcasm, anger, etc.); and as \textbf{neutral} when there are no emotional or lexical cues to indicate the speaker's sentiment one way or another.  If the utterance is too short to determine the sentiment, we marked it as neutral. If the utterance contains a clear mix of positive and negative sentiments, we labeled it as \textbf{mixed}. 

For the test set, each utterance was labeled by 5 passes (from a pool of six annotators) and the segments where at least three annotators agree were retained as ground truth labels. 
We labeled the remaining utterances as \textbf{disagreement}. The inter-annotator Krippendorff's alpha coefficient for this set is 0.48 (considering the data as ordinal). Also, we compute the pairwise Cohen's kappa score for each combination of annotators, getting a mean of 0.37 (lowest:  0.20, highest: 0.47) -- results which are in the range of previous tasks related to natural speech sentiment or emotion
\cite{busso2008iemocap, chen2020large}. For both ASR and Sentiment evaluation on this data, we exclude the samples corresponding to either disagreement or mixed classes.
\kl{are these samples discarded before or after measuring inter-annotator agreement?}\pb{samples were removed after measuring agreemtn, I've seen some papers reporting the agreement before and after the filtering. I prefer before, but let me know what you think}

The fine-tune and dev sets were annotated in a single pass. We leave the decision of using samples labeled as mixed to the user. For our baseline systems, we discard them. Tab.~\ref{tab:voxceleb_stats} shows the distribution of labels obtained for each portion of the dataset.

We also run an independent pass of annotation for approximating human performance for this task on the test set. F1 scores are 0.39 for the negative class, 0.82 for the neutral class, and 0.67 for the positive class.




\vspace{-2mm}
\vspace{-2mm}
\section{Baseline \kledit{models and results}}
\label{sec:baselines}
\noindent We use SLUE-VoxCeleb for ASR and SA and SLUE-VoxPopuli for ASR and NER. Our baselines use pre-trained speech models from the official Fairseq repository.\footnote{\href{https://github.com/pytorch/fairseq}{https://github.com/pytorch/fairseq}} Below we refer to wav2vec 2.0~\cite{baevski2020wav2vec} as W2V2, and refer to all \textit{base} and \textit{large} models with suffix ``-B'' and ``-L'' respectively.  We also refer to the type and amount of unlabeled pretraining data with a suffix on the model name:  LibriSpeech\apremove{ 960 hours} (LS), LibriLight (LL), and multilingual VoxPopuli (VP).
{\it NLP topline} refers to a model that uses human transcription as input, \kledit{which provides the performance of a pipeline system with a perfect ASR model.}
{\it Pipeline} models 
\klremove{use a two stage process with ASR to} generate text from speech \kledit{using an ASR system, and then apply an NLP model on the decoded text.} \klremove{to evaluate the decoded text data
for the downstream task.} \kledit{\it End-to-end (E2E)} models \kledit{directly map from the input speech to output task labels.} \klremove{use speech as input and are directly fine-tuned for the downstream task, without any intermediate procedures as in the pipeline models}
We fine-tune all trainable parameters using the respective \kledit{fine-tune} sets described in Sec.~\ref{sec:annotation}.
\vspace{-2mm}
\subsection{\apedit{Automatic speech recognition}}
\vspace{-1mm}
\label{sec:asr_baseline}
\begin{table}[]
\centering
\resizebox{1.0\linewidth}{!}{%
\begin{tabular}{l|c|c|c|c}
\hlineB{2}
\multirow{2}{*}{Model} & \multicolumn{2}{c|}{VoxPopuli} & \multicolumn{2}{c}{VoxCeleb} \\ \cline{2-5}
 & Dev & Test & Dev & Test \\ \hlineB{2}
W2V2-B-LS960 & 17.5 & 18.4 & 17.1 & 20.9 \\
W2V2-B-VP100K & 21.9 & 22.8 & 29.9 & 33.7 \\
HuBERT-B-LS960 & 18.6 & 19.6 & 20.0 & 21.7 \\
W2V2-L-LL60K & 11.9 & 12.1 & 11.0 & 13.8 \\
\hline
W2V2-B-LS960 (+ in-domain LM) & 14.6 & 15.2 & 15.7 & 18.1 \\
W2V2-B-LS960 (+ TED-LIUM 3 LM) & 12.0 & 12.3 & 13.3 & 16.1 \\
W2V2-L-LL60K (+ in-domain LM) & 12.0 & 12.5 & 12.3 & 13.9 \\
W2V2-L-LL60K (+ TED-LIUM 3 LM) & 9.1 & 9.3 & 9.1 & 11.1 \\
\hlineB{2}
\end{tabular}%
}
\vspace{-2mm}
\caption{ASR performance (WER, \%).
Note that W2V2-B-LS960 and W2V2-L-LL60K get WER 17.6\% and 10.0\%, respectively, on LibriSpeech test-other when fine-tuned on LibriSpeech 10h.
}
\vspace{-1mm}
\label{tab:asr_task_detail}
\end{table}
\textbf{Setup.}
For ASR baselines, we add a linear layer on top of pre-trained W2V2~\cite{baevski2020wav2vec} and HuBERT~\cite{hsu2021hubert} models and fine-tune them on \kledit{the appropriate} SLUE fine-tune set with \kledit{a} character-level CTC objective~\cite{graves2006connectionist}.
When decoding with \kledit{a} language model \apedit{(LM)}, we use beam size 500, LM weight 2 and word insertion penalty -1. \kl{is "word score" the same as "word insertion penalty"?}\ap{yes}\kl{do readers know that?  we could just use "word insertion penalty"}\ap{Sure, done.}
We consider \klremove{two types of LM models:} \kledit{both} in-domain and out-of-domain \kledit{LMs. In-domain LMs are trigram models trained on the SLUE fine-tune set.  Out-of-domain LMs are trigram models trained on the TED-LIUM 3~\cite{hernandez2018ted} LM corpus.\footnote{We found this LM to (slightly) outperform bigram and 4-gram models, as well as LMs trained on LibriSpeech.}}

\noindent\textbf{\kledit{Results.}}
Tab.~\ref{tab:asr_task_detail} shows the ASR performance of baseline models.
W2V2-L-LL60K performs the best (14.3\% and 15.6\% on VP and VC respectively) and W2V2-B-LS960 outperforms other base size models.
Decoding with LM reduces the WER significantly (14.3\% $\rightarrow$ 10.4\% and 15.6\% $\rightarrow$ 12.0\%).
Given that the in-domain transcription is limited, we observe a larger gain in performance when using LM trained on TED-LIUM 3.  \kledit{We note that these WERs are roughly in line with the WERs of the same models on the LibriSpeech test-other set when fine-tuned on 10h of LibriSpeech, making these sets roughly similar in difficulty.  We also note that W2V2-B-LS960 outperforms W2V2-B-VP100K, indicating that the model pre-trained on same-language data is a better fit than the model pre-trained on same-domain data.} \kl{added these sentences for context.  We could also move the LS test-other WERs from the Table 4 caption to this par.}

\vspace{-2mm}
\subsection{\apremove{NER}\apedit{Named entity recognition}}
\vspace{-1mm}

\label{sec:ner_baseline}
\begin{table}[]
\centering
\tablestyle{2pt}{1.1}
\begin{tabular}{lcc|cc}
\hlineB{2}
\multicolumn{1}{c}{Speech model} & LM & Text model & F1 (\%) & label-F1 (\%)  \\
\hlineB{2} 
\textbf{NLP Toplines:}  & & & &  \\
\hspace{3mm} \multirow{1}{*}{ N/A (GT Text)} & \multirow{1}{*}{ N/A} & DeBERTa-L  & 81.4 & 85.7 \\
\hlineB{2}
\textbf{Pipeline approaches:}  & & & &  \\
\hspace{2mm} W2V2-B-LS960 & - & DeBERTa-L & 49.5 & 74.2 \\
\hspace{2mm} W2V2-L-LL60K & - & DeBERTa-L & 59.7 & 78.8 \\
\hspace{2mm} W2V2-B-LS960 & \cmark & DeBERTa-L & 69.2 & 79.8 \\
\hspace{2mm} W2V2-L-LL60K & \cmark & DeBERTa-L & 71.8 & 82.2 \\
\hlineB{2}
\textbf{E2E approaches:}  & & & &  \\
\hspace{2mm} W2V2-B-LS960 & - & \multirow{8}{*}{\normalsize N/A} & 49.6 & 64.0 \\
\hspace{2mm} W2V2-B-VP100K & - &  & 47.9 & 60.8 \\
\hspace{2mm} HuBERT-B-LS960 & - &  & 49.8 & 62.9 \\
\hspace{2mm} W2V2-L-LL60K & - &  & 50.5 & 64.9 \\
\hspace{2mm} W2V2-B-LS960 & \cmark &  & 63.4 & 71.7 \\
\hspace{2mm} W2V2-B-VP100K & \cmark &  & 61.8 & 69.8 \\
\hspace{2mm} HuBERT-B-LS960 & \cmark &  & 61.9 & 70.3 \\
\hspace{2mm} W2V2-L-LL60K & \cmark &  & 64.8 & 73.4 \\
\hlineB{2}
\end{tabular}
\vspace{-2mm}
\caption{Named entity recognition performance\apedit{ on test set}.}
\vspace{-4mm}
\label{tab:ner_task_detail}
\end{table}

\textbf{Setup.}
The NER annotations have the text transcripts along with the (entity phrase, entity tag) pairs. We formulate E2E NER as character level prediction where entity phrases are delimited by tag-specific special characters as in~\cite{ghannay2018end, yadav2020end}. For example, the phrases ``irish" and ``eu" are tagged as NORP (\textcolor{blue}{\$}) and GPE (\textcolor{blue}{\%}) respectively in `\textit{`the \textcolor{blue}{\$ irish ]} system works within a legal and regulatory policy directive framework dictated by the \textcolor{blue}{\% eu ]}}". The vocabulary includes 19 special characters, 18 for each of the entity tags (Tab.~\ref{tab:voxpopuli_stats}), inserted at the beginning of an entity phrase, and one to denote end of the entity phrase. 
The NER E2E baseline models using pre-trained W2V2 and HuBERT models are trained similarly to the ASR models (Sec.~\ref{sec:asr_baseline}).
For \apedit{E2E} experiments with LM, we use a 4-gram LM trained on the SLUE-VoxPopuli fine-tune set, and use decoding parameters tuned to optimize the dev performance of the E2E W2V2-B-LS960 model (beam size 500, LM weight 2 and word insertion penalty 1). \apedit{For pipeline experiments, we use the best ASR models described in Sec.~\ref{sec:asr_baseline}.}\apremove{The LM used in pipeline models is similarly trained and tuned on SLUE-VoxPopuli's ASR fine-tune and dev sets respectively, and use beam size 1000, LM weight 2, and word insertion penalty 0.} 
The pre-trained DeBERTa-L model is fine-tuned for NER (token classification) using HuggingFace's transformers toolkit~\cite{wolf2020transformers} where a linear layer is added on top of the final hidden-state output. 

\noindent\textbf{Results}.
Baseline experiments are reported in Tab.~\ref{tab:ner_task_detail}. 
Note that\kledit{, since the LM was fine-tuned for a single model, the performance of other models } could be sub-optimal; 
similarly the NLP models were not tuned for the Pipeline experiments.
The best models are chosen based on the WER of NER-annotated sentences in \kledit{the} dev set. 
We make the following observations:
\begin{itemize}[leftmargin=*,noitemsep,nolistsep]
    \item There is significant room for improvement for both Pipeline and E2E models, even while leveraging state-of-the-art pre-trained models.
    \item W2V2-B-LS960 and HuBERT-B-LS960 outperform W2V2-B-VP100k suggesting that language mismatch is worse than domain mismatch in a pre-trained model. \apremove{The W2V2-B-LL60K performance is poorer than expected, which is due to the limited training budget for that experiment.}
    \item LM decoding provides consistent and significant improvements. \apremove{even for the sub-optimally trained W2V2-B-LL60K}. 
    \item Improvements from larger speech models are less evident when using LM decoding; that is, using a small amount (5k utterances) of unlabeled text is as beneficial as leveraging 60 times more unlabeled audio data with the current methods.
\end{itemize}
The last point may suggest that the pre-trained speech models do not learn significant semantic information, so that even a small amount of additional
semantic knowledge (in the form of language models here) should help immensely.
This point deserves further exploration in future work.

\vspace{-2mm}
\subsection{\apremove{SA}\apedit{Sentiment analysis}}
\label{sec:sa_baseline}
\begin{table}[]
\centering
\resizebox{1.0\linewidth}{!}{%
\begin{tabular}{lcc|cc}
\hlineB{2}
\multicolumn{1}{c}{Speech model} & LM & Text model & Recall \kledit{ (\%)} & F1 \kledit{ (\%)} \\ \hlineB{2}
\textbf{NLP Toplines} : & & & & \\ 
\hspace{2mm}  N/A (GT Text) & N/A & DeBERTa-L & - & 66.8 \\ \hlineB{2}
\textbf{Pipeline approaches} : & & & & \\ 
\hspace{2mm} W2V2-B-LS960 & - & DeBERTa-L & - & 63.6 \\
\hspace{2mm} W2V2-L-LL60K & - & DeBERTa-L & - & 65.7 \\
\hspace{2mm} W2V2-B-LS960 & \cmark & DeBERTa-L & - & 65.4 \\ \hspace{2mm} W2V2-L-LL60K & \cmark & DeBERTa-L & - & 65.5 \\ \hlineB{2}
\textbf{E2E approaches} : & & & & \\ 
\hspace{2mm} W2V2-B-LS960 & \multirow{4}{*}{N/A}  & \multirow{4}{*}{N/A} & - & 48.6 \\ 
\hspace{2mm} W2V2-B-VP100K & & & - & 38.9  \\ 
\hspace{2mm} HuBERT-B-LS960 & & & - & 49.4 \\ 
\hspace{2mm} W2V2-L-LL60K & & & - & 50.1 \\ \hlineB{2}
\end{tabular}%
}
\vspace{-2mm}
\caption{Sentiment \kledit{a}nalysis performance\apedit{ on test set}\kledit{.}\klremove{ in unweighted Recall and F1 score}\kl{shortened caption for space and to match the NER table, and realigned "speech model" to match table 5} }
\vspace{-5mm}
\label{tab:sa_task_detail}
\end{table}

\vspace{-1mm}

\textbf{Setup.}
For \kledit{the} NLP topline, we use\klremove{d} \kledit{the ground-truth (GT)} text of \kledit{the} fine-tune set for fine-tuning the NLP model (DeBERTa-L) and use\klremove{d} \kledit{the} GT text of \kledit{the} dev and test set\kledit{s} for evaluation. For pipeline \kledit{models}, we use\klremove{d} \kledit{the} GT text of \kledit{the} fine-tune set for fine-tuning the NLP models and \kledit{the} ASR transcription\kledit{s} of \kledit{the} dev and test set\kledit{s} for evaluation. For the ASR-generated transcription, we use\klremove{d} the \suwonadd{best} ASR model\kledit{s} described in Sec. \ref{sec:asr_baseline}. For E2E \kledit{models}, we use\klremove{d} the speech waveform\kledit{s} of \kledit{the} fine-tune set for fine-tuning the speech models (W2V2 and HuBERT) and \kledit{the} speech waveform\kledit{s} of \kledit{the} dev and test \kledit{sets} for evaluation. \kl{could remove the prev sentence -- I think it's clear what we do with E2E models without stating it} \kledit{For E2E models, o}n top of the pre-trained speech model, we add\klremove{ed} \kledit{a} self\kledit{-}attention pooling layer and 2 fully connected layers including \kledit{the} output layer.

\noindent\textbf{Result\kledit{s}.}
A detailed baseline performance evaluation is shown in Tab.~\ref{tab:sa_task_detail}. 
In \kledit{the} pipeline experiments, \kledit{a} larger speech model\kledit{ produces} no notable gain in \kledit{F1} score although it \kledit{improves the ASR WER by 5\% absolute.} Similarly, LM decoding in pipeline \kledit{models} doesn't help \kledit{on the} SA task.
More interestingly, the pipeline system performs almost \kledit{as well as} the NLP topline system, \kledit{suggesting that the} NLP model can tolerate \kledit{ASR errors} for \kledit{the} SA task.
The E2E systems \kledit{have much worse performance than} the pipeline approaches. \suwonadd{The performance of E2E system heavily depends on what pre-trained speech model is used while pipeline systems is not sensitive to speech model.}
\kl{I don't quite follow the last 2 sentences}\suwon{updated.}\kl{still not completely following.  Besides W2V2-B-VP100K, the E2E approaches have a couple \% differences depending on the pre-trained model, which doesn't seem so heavily dependent as we claim.  }
\vspace{-2mm}
\subsection{SLUE benchmark results}
\vspace{-1mm}

\label{sec:results}

\begin{table}[]
\centering
\resizebox{\linewidth}{!}{%
\tablestyle{1.5pt}{1.0}
\begin{tabular}{lcc|c|c|c|c|c}
\hlineB{2}
\multicolumn{3}{c|}{Model} & \multirow{2}{*}{\begin{tabular}[c]{@{}c@{}}SLUE\\ Score\end{tabular}} & \multicolumn{2}{c|}{ASR} & NER & SA \\ \cline{1-3} \cline{5-8}
Speech Model  & LM & Text &                                                                       & VC         & VP         & VP  & VC \\
\hlineB{2}
\textbf{NLP Topline}: & & & & & & & \\
\hspace{1mm} N/A (GT Text) & N/A & DeBERTa-L & 82.7 & 0.0 & 0.0 & 81.4 & 66.8 \\
\hlineB{2}
\textbf{Pipeline}: & & & & & & \\
\hspace{1mm} W2V2-L-LL60K & - & DeBERTa-L & 70.8 & 12.1 & 13.8 & 59.7 & 65.7 \\
\hspace{1mm} W2V2-L-LL60K & \cmark & DeBERTa-L & 75.7 & 9.3 & 11.1 & 71.8 & 65.5 \\
\hlineB{2}
\textbf{E2E}: & & \multirow{6}{*}{\normalsize N/A} & & & & &\\
\hspace{1mm} W2V2-B-LS960 & - &  & 59.4 & 18.4 & 20.6 & 49.6 & 48.1 \\
\hspace{1mm} HuBERT-B-LS960 & - &  & 59.5 & 19.6 & 21.5 & 49.8 & 49.2 \\
\hspace{1mm} W2V2-L-LL60K & - &  & 62.5 & 12.1 & 13.5 & 50.5 & 49.8 \\
\hspace{1mm} W2V2-B-LS960 & \cmark* &  & 65.8 & 12.3 & 15.8 & 63.4 & 48.1 \\
\hspace{1mm} W2V2-L-LL60K & \cmark* &  & 68.2 & 9.3 & 10.9 & 64.8 & 49.8 \\
 \hlineB{2}
\end{tabular}%
    }
\caption{Performance of baseline models\apedit{ on test set}\apedit{; VP: VoxPopuli, VC: VoxCeleb}. \kl{need to increase the font size.  Maybe remove the "(\%)"s, and indent the model names less?  not sure if that would help enough but it's a start}
\apedit{(\% WER for ASR and \% F1-scores for NER and SA).}
*: used LM decoding for ASR and NER tasks.
}
\vspace{-5mm}
\label{tab:slue_performance}
\end{table}

A summary of the main SLUE benchmark results thus far is shown in Tab.~\ref{tab:slue_performance}. As an overall rating, we define the SLUE benchmark score as an average of the primary metrics for the three tasks: $\mathrm{SLUE}_{\text{score}} = \frac{1}{3}\left((100 - \frac{\mathrm{WER}_\text{ASR-VP} + \mathrm{WER}_\text{ASR-VC}}{2}) + \mathrm{F1}_\text{NER-VP} + \mathrm{F1}_\text{SA-VC}\right)$.
The best pipeline system is still 11\% behind the NLP topline in terms of SLUE score.
Compared to the best pipeline system, the best E2E system is 7\% behind, likely because of the huge NLP model size in the pipeline systems.
Adding a language model improves both pipeline and E2E approaches, especially for the NER task.

\vspace{-1mm}
\section{Conclusions}
\vspace{-2mm}

\label{sec:conclusions}
Motivated by the growing interest in SLU tasks and recent progress on pre-trained \klremove{generic }representations, we have proposed a new benchmark suite consisting of newly annotated fine-tuning and evaluation sets, and have provided annotations and baselines for new NER, sentiment, and ASR evaluations.
We have presented the detailed process to build the benchmark and evaluated numerous
baseline systems using current state-of-the-art speech and NLP models. Future work includes adding more SLU tasks and refining test set annotation and human performance measures.
Additional potential future work includes more speech-specific evaluation, such as measuring the accuracy of named entity time boundaries in the NER task.







\clearpage

\clearpage
\vfill\pagebreak

\bibliographystyle{IEEEbib}
{\footnotesize \bibliography{refs}}

\ifarxiv
\onecolumn 
\normalsize

\section{Appendix}

\textbf{A. Detailed NER label statistics in addition to Sec.~\ref{sec:annotation}.1}

\begin{table}[h]
\centering
\resizebox{0.6\linewidth}{!}{%
\begin{tabular}{l|l|l|l}
\hlineB{2}
\begin{tabular}[c]{@{}c@{}}Combined\\ label\end{tabular} & \begin{tabular}[c]{@{}c@{}}Raw label\\ (ontonotes5)\end{tabular} & \begin{tabular}[c]{@{}c@{}}\# of NER phrases\\ (fine-tune/dev/test)\end{tabular} & \begin{tabular}[c]{@{}c@{}}\# of distinct NER phrases\\ (fine-tune/dev/test)\end{tabular} \\\hlineB{2}
\multirow{2}{*}{PLACE} & GPE &1641 / 500 / 560 &162 / 96 / 120 \\
& LOC &371 / 142 / 171 &56 / 27 / 34 \\
\hline
\multirow{5}{*}{QUANT} & CARDINAL &584 / 193 / 171 &137 / 68 / 84 \\
& ORDINAL &267 / 110 / 57 &19 / 14 / 10 \\
& MONEY &60 / 18 / 8 &52 / 16 / 6 \\
& PERCENT &3 / 3 / 2 &3 / 2 / 2 \\
& QUANTITY &9 / 3 / 8 &9 / 2 / 8 \\
\hline
\multirow{1}{*}{ORG} & ORG &864 / 259 / 273 &255 / 100 / 126 \\
\hline
\multirow{2}{*}{WHEN} & DATE &723 / 259 / 179 &327 / 154 / 119 \\
& TIME &39 / 1 / 7 &22 / 1 / 6 \\
\hline
\multirow{1}{*}{NORP} & NORP &647 / 220 / 348 &128 / 60 / 91 \\
\hline
\multirow{1}{*}{PERSON} & PERSON &272 / 51 / 81 &201 / 44 / 59 \\
\hline
\multirow{1}{*}{LAW} & LAW &250 / 60 / 96 &156 / 41 / 75 \\
\hline
\multirow{5}{*}{Discarded tags} & EVENT &73 / 37 / 21 &49 / 23 / 19 \\
& FAC &10 / 2 / 4 &9 / 2 / 4 \\
& LANGUAGE &2 / 0 / 11 &1 / 0 / 7 \\
& PRODUCT &2 / 3 / 6 &2 / 2 / 6 \\
& WORK OF ART &3 / 1 / 3 &3 / 1 / 3 \\
\hline
\end{tabular}%
}
\caption{SLUE-VoxPopuli NER label statistics}
\label{tab:voxpopuli_stats_detail}
\end{table}

\noindent\textbf{B. SLUE-VoxPopuli test set NER annotation disagreement analysis in addition to Sec.~\ref{sec:annotation}.1}

\begin{figure}[h]
    \centering
    \includegraphics[width=0.5\linewidth]{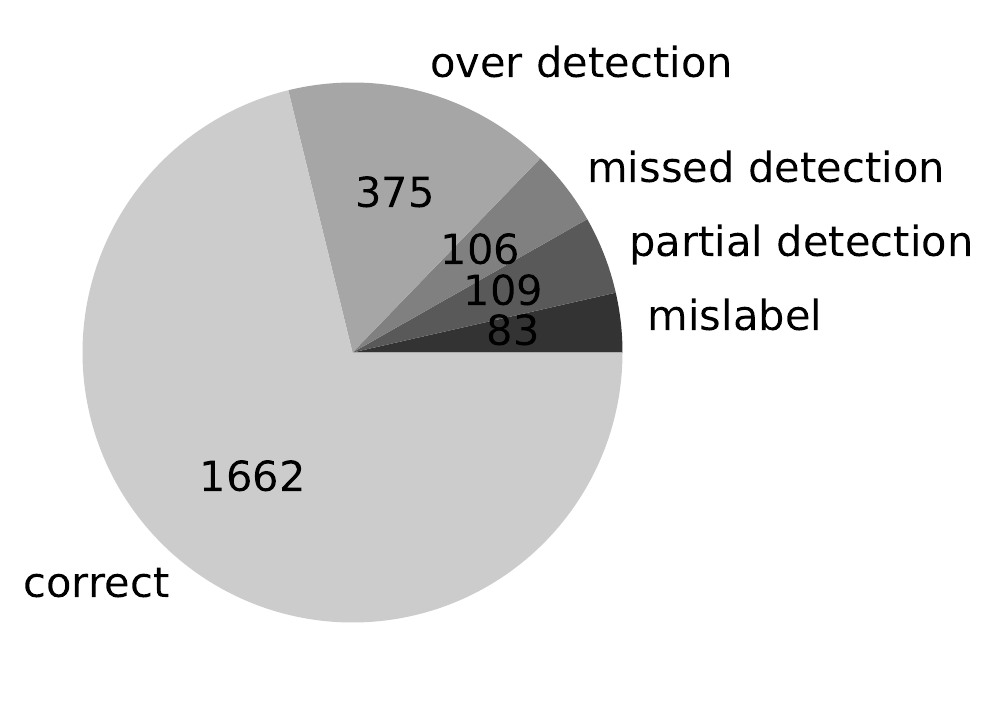}
    \caption{Classification of disagreements between the two annotation passes on SLUE-VoxPopuli test set. \kl{there seems to be a stray "0.5" below "correct", or am I imagining?}\ap{fixed}
    }
    \label{fig:ner_pie}
\end{figure}

\clearpage
\noindent\textbf{C. SLUE-VoxCeleb Test set inter-pass analysis in addition to Sec.~\ref{sec:annotation}.2}
\begin{figure}[h]
    \centering
    \includegraphics[width=0.5\linewidth]{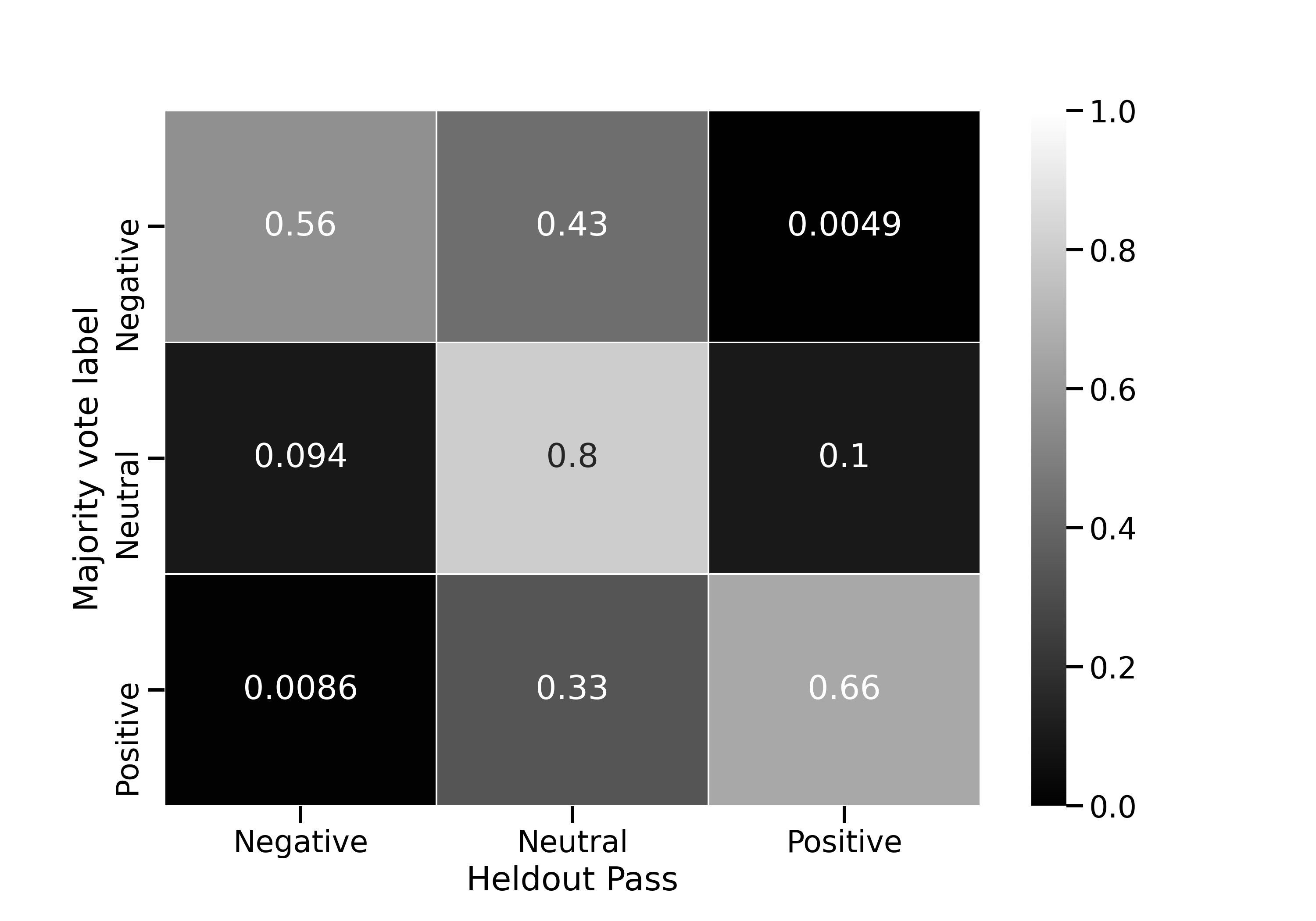}
    \caption{Confusion matrix between Majority vote and heldout pass (4th pass)}
    \label{fig:my_label}
\end{figure}

\noindent\textbf{D. Detailed SLUE Benchmark v0.2 result on Dev set}
\begin{table}[h]
\centering
\resizebox{0.65\linewidth}{!}{%
\begin{tabular}{lcc|c|c|c|c|c}
\hlineB{2}
\multicolumn{3}{c|}{\multirow{2}{*}{Model}} & \multirow{3}{*}{\begin{tabular}[c]{@{}c@{}}SLUE \\Score\end{tabular}} & \multicolumn{2}{c|}{ASR} & NER & SA \\ \cline{5-8}
\multicolumn{3}{c|}{}  & &\multicolumn{2}{c|}{WER(\%) } & F1 (\%) & F1(\%) \\ \cline{1-3} \cline{5-8}
Speech \kledit{model}  & LM & Text &  & VP & VC & VP & VC \\
\hline
\textbf{NLP Topline}: & & & & & & & \\
\hspace{1mm} \multirow{3}{*}{\begin{tabular}[c]{@{}c@{}}N/A\\(GT text)\end{tabular}} & \multirow{3}{*}{N/A} & BERT-B & . & 0.0 & 0.0 & 86.2 & 53.7 \\
\hspace{1mm}  & & DeBERTa-B & . & 0.0 & 0.0 & 86.0 & 57.2 \\
\hspace{1mm}  & & DeBERTa-L & . & 0.0 & 0.0 & 87.5 & 57.8 \\
\hline
\textbf{Pipeline}: & & & & & & & \\
\hspace{1mm} W2V2-B-LS960 & - & BERT-B & 62.1 &    17.5 &    17.5 &     52.0 &    51.9 \\
\hspace{1mm} W2V2-B-LS960 & - & DeBERTa-B & 61.7 &    17.5 &    17.5 &     49.9 &    52.6 \\
\hspace{1mm} W2V2-B-LS960 & - & DeBERTa-L & 64.4 &    17.5 &    17.5 &     55.2 &    55.7 \\
\hspace{1mm} W2V2-L-LL60K & - & DeBERTa-L & 69.8 &    11.9 &    11.3 &     65.0 &    56.1 \\
\hspace{1mm} W2V2-B-LS960 & \cmark & BERT-B & 71.6 &    12.0 &    13.3 &     73.0 &    54.4 \\
\hspace{1mm} W2V2-B-LS960 & \cmark & DeBERTa-B & 70.2 &    12.0 &    13.3 &     72.0 &    51.3 \\
\hspace{1mm} W2V2-B-LS960 & \cmark & DeBERTa-L & 72.0 &    12.0 &    13.3 &     73.8 &    54.9 \\
\hspace{1mm} W2V2-L-LL60K & \cmark & DeBERTa-L & 74.6 &     9.1 &     9.1 &     76.7 &    56.3 \\
\hline
\textbf{E2E}: & & & & & & & \\
\hspace{1mm} W2V2-B-LS960 & - &  \multirow{8}{*}{N/A} & 60.3 &    17.5 &    17.5 &     55.0 &    43.3 \\
\hspace{1mm} W2V2-B-VP100K & - &  & 55.0 &    21.9 &    30.5 &     53.2 &    38.0 \\
\hspace{1mm} HuBERT-B-LS960 & - &  &  60.2 &    18.6 &    20.0 &     54.5 &    44.0 \\
\hspace{1mm} W2V2-L-LL60K & - &  &  63.4 &    11.9 &    11.3 &     56.6 &    45.3 \\
\hspace{1mm} W2V2-B-LS960 & \cmark* &  & 66.2 &    12.0 &    13.3 &     68.1 &    43.3 \\
\hspace{1mm} W2V2-B-VP100K & \cmark* &  & 62.3 &    16.8 &    19.7 &     67.3 &    38.0 \\
\hspace{1mm} HuBERT-B-LS960 & \cmark* &  & 65.4 &    15.9 &    15.2 &     67.8 &    44.0 \\
\hspace{1mm} W2V2-L-LL60K & \cmark* &  & 68.8 &     9.1 &     9.1 &     70.3 &    45.3 \\
 \hlineB{2}
\end{tabular}%
    }
\caption{Performance of baseline models.
*: used LM decoding for ASR and NER tasks.
}
\vspace{-5mm}
\label{tab:slue_performance_detail}
\end{table}

\newpage
\noindent\textbf{E. Detailed SLUE Benchmark v0.2 result on Test set in addition to Sec.~\ref{sec:results}}
\begin{table}[h]
\centering
\resizebox{0.65\linewidth}{!}{%
\begin{tabular}{lcc|c|c|c|c|c}
\hlineB{2}
\multicolumn{3}{c|}{\multirow{2}{*}{Model}} & \multirow{3}{*}{\begin{tabular}[c]{@{}c@{}}SLUE \\Score\end{tabular}} & \multicolumn{2}{c|}{ASR} & NER & SA \\ \cline{5-8}
\multicolumn{3}{c|}{}  & &\multicolumn{2}{c|}{WER(\%) } & F1 (\%) & F1(\%) \\ \cline{1-3} \cline{5-8}
Speech \kledit{model}  & LM & Text &  & VP & VC & VP & VC \\
\hline
\textbf{NLP Topline}: & & & & & & & \\
\hspace{1mm} \multirow{3}{*}{\begin{tabular}[c]{@{}c@{}}N/A\\(GT text)\end{tabular}} & \multirow{3}{*}{N/A} & BERT-B & 81.5 & 0.0 & 0.0 & 81.2 & 63.3 \\
\hspace{1mm}  & & DeBERTa-B & 82.3 & 0.0 & 0.0 & 81.4 & 65.5 \\
\hspace{1mm}  & & DeBERTa-L & 82.7 & 0.0 & 0.0 & 81.4 & 66.8 \\
\hline
\textbf{Pipeline}: & & & & & & & \\
\hspace{1mm} W2V2-B-LS960 & - & BERT-B & 62.6 &    18.4 &    20.9 &     47.4 &    60.1 \\
\hspace{1mm} W2V2-B-LS960 & - & DeBERTa-B & 63.0 &    18.4 &    20.9 &     46.4 &    62.1 \\
\hspace{1mm} W2V2-B-LS960 & - & DeBERTa-L & 64.5 &    18.4 &    20.9 &     49.5 &    63.6 \\
\hspace{1mm} W2V2-L-LL60K & - & DeBERTa-L & 70.8 &    12.1 &    13.8 &     59.7 &    65.7 \\
\hspace{1mm} W2V2-B-LS960 & \cmark & BERT-B & 71.6 &    12.3 &    16.1 &     68.6 &    60.5 \\
\hspace{1mm} W2V2-B-LS960 & \cmark & DeBERTa-B & 72.6 &    12.3 &    16.1 &     68.5 &    63.6 \\
\hspace{1mm} W2V2-B-LS960 & \cmark & DeBERTa-L & 73.4 &    12.3 &    16.1 &     69.2 &    65.4 \\
\hspace{1mm} W2V2-L-LL60K & \cmark & DeBERTa-L & 75.7 &     9.3 &    11.1 &     71.8 &    65.5 \\
\hline
\textbf{E2E}: & & & & & & & \\
\hspace{1mm} W2V2-B-LS960 & - &  \multirow{8}{*}{N/A} & 59.5 &    18.4 &    20.9 &     49.6 &    48.6 \\
\hspace{1mm} W2V2-B-VP100K & - &  & 52.8 &    22.8 &    33.7 &     47.9 &    38.9 \\
\hspace{1mm} HuBERT-B-LS960 & - &  & 59.5 &    19.6 &    21.7 &     49.8 &    49.4 \\
\hspace{1mm} W2V2-L-LL60K & - &  & 62.5 &    12.1 &    13.8 &     50.5 &    50.1 \\
\hspace{1mm} W2V2-B-LS960 & \cmark* &  & 65.9 &    12.3 &    16.1 &     63.4 &    48.6 \\
\hspace{1mm} W2V2-B-VP100K & \cmark* &  & 60.2 &    17.3 &    23.0 &     61.8 &    38.9 \\
\hspace{1mm} HuBERT-B-LS960 & \cmark* &  & 64.9 &    16.8 &    16.9 &     61.9 &    49.4 \\
\hspace{1mm} W2V2-L-LL60K & \cmark* &  & 68.2 &     9.3 &    11.1 &     64.8 &    50.1 \\
 \hlineB{2}
\end{tabular}%
    }
\caption{Performance of baseline models.
*: used LM decoding for ASR and NER tasks.
}
\vspace{-5mm}
\label{tab:slue_performance_detail}
\end{table}
\fi

\end{document}